\documentclass{l4dc2026}
\jmlrvolume{}
\jmlryear{}
\jmlrproceedings{}{}
\jmlrpages{}  
\usepackage[font=small,labelfont=bf]{caption} 
\usepackage{caption}
\usepackage{graphicx}

\usepackage{amsmath}
\usepackage{amssymb}
\usepackage{amsfonts}
\usepackage{mathtools}
\usepackage{MnSymbol}
\usepackage{algorithm}
\usepackage{algpseudocode}
\usepackage{tikz}
\usepackage{xcolor}
\usepackage{enumitem}
\usepackage[usenames,dvipsnames,svgnames,table]{xcolor}
\usepackage{float}
\floatstyle{plaintop}
\restylefloat{table}
\usepackage{cite}
\usepackage{wrapfig}

\usepackage[normalem]{ulem}

\newtheorem{problem}{Problem}

\title[]{Time-aware Motion Planning in Dynamic Environments with Conformal Prediction}
\usepackage{times}

\author{%
 \Name{Kaier Liang$^{1}$} \Email{kal221@lehigh.edu}
 \AND
 \Name{Licheng Luo$^{2}$} \Email{lichengl@ucr.edu}
 \AND
 \Name{Yixuan Wang$^{2}$} \Email{ywang1457@ucr.edu}
 \AND
 \Name{Mingyu Cai$^{2}$} \Email{mingyu.cai@ucr.edu}
 \AND
 \Name{Cristian Ioan Vasile$^{1}$} \Email{cvasile@lehigh.edu}\\
 \addr $^{1}$Mechanical Engineering and Mechanics Department, Lehigh University, PA, USA\\
 \addr $^{2}$Department of Mechanical Engineering, University of California Riverside, CA, USA
}

\begin{document}

\maketitle

\begin{abstract}%
Safe navigation in dynamic environments remains challenging due to uncertain obstacle behaviors and the lack of formal prediction guarantees. We propose two motion planning frameworks that leverage conformal prediction (CP): a global planner that integrates Safe Interval Path Planning (SIPP) for uncertainty-aware trajectory generation, and a local planner that performs online reactive planning. The global planner offers distribution-free safety guarantees for long-horizon navigation, while the local planner mitigates inaccuracies in obstacle trajectory predictions through adaptive CP, enabling robust and responsive motion in dynamic environments.
To further enhance trajectory feasibility, we introduce an adaptive quantile mechanism in the CP-based uncertainty quantification. Instead of using a fixed confidence level, the quantile is automatically tuned to the optimal value that preserves trajectory feasibility, allowing the planner to adaptively tighten safety margins in regions with higher uncertainty.
We validate the proposed framework through numerical experiments conducted in dynamic and cluttered environments. 
The project page is available at \url{https://time-aware-planning.github.io}

\end{abstract}


\section{Introduction}
Robust navigation in dynamic environments remains a central challenge for autonomous agents~\citet{mavrogiannis2023core,rudenko2020human}. Applications such as mobile robots navigating crowded areas or drones operating in shared airspace require planners that can efficiently compute collision-free trajectories despite the uncertainty in predicting the motion of surrounding obstacles. Classical approaches to motion planning, such as dynamic window search or sampling-based planners~\citet{aoude2013probabilistically}, either neglect prediction uncertainty or rely on strong parametric assumptions about obstacle dynamics, resulting in overly conservative behavior or unsafe plans~\citet{phillips2011sipp}.
To handle such uncertainty, a variety of robust planning techniques have been developed. Chance-constrained motion planning~\citet{blackmore2011chance,du2011robot} incorporates stochastic bounds on obstacle trajectories, ensuring constraint satisfaction with high probability. Distributionally robust formulations extend this idea by optimizing against worst-case distributions within ambiguity sets—recent robotics examples include Wasserstein-robust risk maps and motion/control schemes~\citet{hakobyan2021wasserstein}. Tube-based MPC contains deviations within tightened constraint sets and provides robust feasibility under bounded disturbances~\citet{zhang2022robust,mayne2011tube}.
Reference governor-based methods~\citep{garone2017reference,liang2024control} address constraint satisfaction in complex systems by computing safety margins and navigation fields that guide the system toward feasible operating regions.
Risk-sensitive /Conditional Value at Risk (CVaR)-based planners explicitly trade expected cost against tail risk~\citep{chow2015risk}. 
These methods provide theoretical safety guarantee but typically require strong assumptions on the underlying noise model (e.g., Gaussian errors, convex uncertainty sets) and often scale poorly in high-dimensional or long-horizon planning tasks. 
Modern online control approaches leverage Control Barrier Functions (CBFs)~\citep{ames2019control,lopez2020robust} to ensure safety under uncertainty, incorporating robust and adaptive techniques to compensate for noisy observations and disturbance.

Recent advances in trajectory prediction using deep learning models provide high-quality forecasts of dynamic agents, but they lack calibrated measures of uncertainty. As a result, plans based directly on such predictions risk unsafe outcomes when the predictions deviate from reality. Conformal prediction (CP) has emerged as a powerful framework for providing distribution-free uncertainty quantification with finite-sample guarantees~\citep{vovk2005algorithmic, sun2023conformal,strawn2023conformal,chee2024uncertainty,liang2025safe}. By producing valid prediction regions around obstacle trajectories at a user-specified confidence level, CP enables principled reasoning about probabilistic guarantee.
It has been applied to probabilistic verification tasks such as large language model validation~\citep{wang2024conformal,cherian2024large}, temporal logic verification~\citep{yu2026signal}, semantic segmentation~\citep{mossina2024conformal} and so on.
Lindemann et al. demonstrated how CP can be integrated into a model predictive control (MPC) framework to provide probabilistic safety guarantee in continuous state spaces~\citep{lindemann2023safe}. While this represents an important step forward, MPC-based formulations can be computationally expensive for long horizons, and their continuous nature makes them less suited for discrete, graph-based planning domains such as navigation on road networks, grids, or task-specific graphs. While split conformal prediction guarantee finite sample coverage under exchangeability, real deployments rarely preserve exchangeability between historical calibration data amd the incoming stream. Recent adaptive conformal prediction(ACP) methods address this gap by coupling an online exchangeability diagnostic with a feedback-driven update that steers realized miscoverage to a target level without assuming exchangeability.~\citep{dixit2022adaptiveconformalpredictionmotion, gibbs2021adaptiveconformalinferencedistribution,JMLR:v25:22-1218}

In this work, we integrate conformal prediction into Safe Interval Path Planning (SIPP)~\citep{phillips2011sipp} and extend it to a sampling-based framework for continuous domains. Our approach augments SIPP with confidence levels derived from conformal prediction, enabling the planner to reason about probabilistic safety alongside travel time. To handle continuous environments where online sensor feedback is available, we further develop a time-aware Adaptive Conformal Prediction Rapidly-Exploring Random Tree (ACP-RRT) algorithm that leverages real-time observations at each step to adaptively calibrate safety bounds, enabling local reactive planning while maintaining distribution-free safety guarantee. Together, these frameworks demonstrate how conformal prediction can provide unified, uncertainty-aware motion planning across both discrete and continuous settings—with Conformal Prediction Safe Interval Path Planning  (CP-SIPP) suited for global, long-horizon planning and ACP-RRT designed for local, reactive navigation in response to incoming sensor data.

The main contributions of this paper are: (i) We introduce a CP-SIPP framework that integrates conformal prediction for global, long-horizon navigation with formal, distribution-free safety guarantee in discrete spatial-temporal domains. (ii) We propose optimization methods that balance trajectory feasibility, optimal cost, and CP confidence.
(iii) We extend the conformal safety principle to continuous domains through a time-aware ACP-RRT that performs local, reactive planning by adaptively calibrating uncertainty bounds using online sensor feedback at each step, enabling probabilistically safe motion planning under distribution shift.


\section{Preliminaries and Problem Formulation}

\subsection{Confidence Prediction for Dynamic Obstacles}\label{sec:cp-basics}
Consider a controllable robot operating on a finite, undirected graph 
$G = (V, E)$, where $V \subset \mathbb{Z}^2$ represents discrete spatial 
locations and $E \subseteq V \times V$ encodes valid transitions between them. 
The agent’s trajectory over a time horizon $T$ is a sequence of vertices 
$\pi = (v_0, v_1, \ldots, v_k)$, where $v_t \in V$ and $k \leq T$.
The environment contains $n$ dynamic obstacles 
$\mathcal{O} = \{\tau_1, \tau_2, \ldots, \tau_n\}$, 
where each obstacle follows a continuous trajectory 
$\tau_i : [0, T] \to \mathbb{R}^2$ specifying its position over time. 
The true obstacle trajectories $\tau_i(t)$ are unknown and must be estimated 
from sensor data and motion models, yielding predicted trajectories 
$\hat{\tau}_i(t)$.

Since the predictions $\hat{\tau}_i(t)$ inevitably deviate from the true 
future positions, we employ \emph{conformal prediction} ~\citep{lei2017distributionfreepredictiveinferenceregression}, 
a statistical framework for constructing prediction regions with 
finite-sample, distribution-free guarantees. 
CP requires only the assumption of \emph{exchangeability} between calibration 
and test data, rather than any specific parametric form, 
making it particularly suitable for real-world settings 
with uncertain or nonstationary dynamics.
The key mechanism of CP is the \emph{nonconformity score}, 
which quantifies the discrepancy between model predictions and ground truth. 
We define the nonconformity score at time $t$ as the maximum 
prediction error across all obstacles, 
$R(t) = \max_{i = 1, \ldots, n} \| \hat{\tau}_i(t) - \tau_i(t) \|$~\citep{lindemann2023safe}. 

Given a calibration dataset $\mathcal{D}_{\text{cal}}$ of historical obstacle 
trajectories, the conformal framework assumes that calibration and deployment 
data are exchangeable. 
When this assumption does not hold (e.g., under time-varying dynamics), 
a \emph{weighted conformal prediction} approach~\citep{barber2023conformal} 
can be applied. 
The prediction threshold $C_t$ is then chosen as the $(1 - \alpha)$-quantile 
of the empirical distribution of calibration nonconformity scores, 
$C_t = \text{Quantile}_{1 - \alpha}\{R_j(t) : j \in \mathcal{D}_{\text{cal}}\}$, 
where $\alpha \in [0, 1]$ is the user-specified miscoverage rate. 
This construction ensures that $P(R(t) \leq C_t) \geq 1 - \alpha$, 
providing a finite-sample coverage guarantee that holds under exchangeability 
without requiring any distributional assumptions on obstacle motion.




To integrate conformal prediction into discrete motion planning, 
we derive spatially indexed confidence values. 
For each grid location $s \in V$ and time $t$, 
we compute its distance to each predicted obstacle position as 
$d_i(s, t) = \| s - \hat{\tau}_i(t) \|$. 
The confidence that location $s$ is safe at time $t$ is then given by 
$    c(s, t) = 
\frac{
    |\{ j \in \mathcal{D}_{\text{cal}} : R_j(t) \leq \min_i d_i(s, t) \}|
}{
    |\mathcal{D}_{\text{cal}}|
}.$
Here, $R_j(t)$ denotes the nonconformity score computed from the $j$-th 
calibration trajectory. 
Intuitively, $c(s,t)$ represents the empirical fraction of calibration samples 
whose maximum prediction error does not exceed the minimum safety margin 
$\min_i d_i(s, t)$. 
By the conformal prediction guarantee, if $c(s,t) \geq 1 - \alpha$, 
then location $s$ is collision-free at time $t$ with probability at least 
$1 - \alpha$ under exchangeability. 
This formulation yields distribution-free, finite-sample safety guarantees 
that hold regardless of the underlying obstacle motion distribution 
and allows the planner to trade off safety and performance by selecting 
appropriate confidence thresholds for each waypoint—higher confidence levels 
correspond to larger safety margins around predicted obstacle positions.




\subsection{Problem Formulation}
In dynamic environments with uncertain obstacle motion, a valid plan must ensure not only 
spatiotemporal feasibility but also probabilistic safety. 
Classical shortest-path formulations minimize travel time while enforcing connectivity, 
yet they cannot explicitly account for uncertainty induced by prediction errors. 
To bridge this gap, we introduce two complementary formulations: a global planning problem that employs pre-calibrated conformal thresholds 
for long-horizon navigation, and a local reactive planning problem
that adapts safety margins online using sensor feedback.
For global planning in the absence of real-time observations, 
the objective is to compute a trajectory from start to goal 
that is both time-efficient and provably safe with respect to prediction uncertainty.

\vspace{-0.2cm}

\begin{problem}[Global Time-aware Motion Planning]
\label{pb:global}
Given a start location $v_{\text{start}}$, a goal location $v_{\text{goal}}$, a time horizon $T$, 
a minimum confidence threshold $c_{\min} \in [0, 1]$, and predicted obstacle trajectories 
$\{\hat{\tau}_i(t)\}_{i=1}^{n}$ with fixed calibration dataset, find an optimal 
trajectory $\pi$ offline that solves:
\begin{subequations}
\label{eq:optimization_problem}
\begin{align}
\min_{\pi} \quad &  \sum_{j=0}^{k-1} \left( \gamma w(v_j, v_{j+1}) + (1- \gamma)c_j \right) \label{eq:objective} \\
\text{subject to} \quad & v_0 = v_{\text{start}}, \quad v_k = v_{\text{goal}} \quad (\exists k \le T) \label{eq:boundary} \\
   & (v_{j}, v_{j+1}) \in E, 
   \quad v_j \in \mathcal{S}(c_j, t_j), \quad \forall j \in \{0, \ldots, k\} \label{eq:safety_constraint}
\end{align}
\end{subequations}
where $\gamma \in [0, 1]$ balances travel time and safety, $w(v_j, v_{j+1})$ is the edge cost e.g., travel time, 
and $\mathcal{S}(c_j, t_j) = \{s \in V : c(s, t_j) \geq c_j\}$ is the conformal safe set at time $t_j$.
\end{problem}

\begin{problem}[Local Reactive Time-aware Motion Planning]
\label{pb:local}
Given current state $(v_{\text{current}}, t_{\text{current}})$, goal region $v_{\text{goal}}$, sensor horizon $H_{\text{local}}$, predicted obstacle trajectories $\{\hat{\tau}_i(t)\}_{i=1}^{n}$, find a feasible trajectory online $\pi_{\text{local}}: [t_{\text{current}}, t_{\text{current}} + H_{\text{local}}] \to \mathcal{X}$ that starts at $v_{\text{current}}$, maintain conformal safety confidence $C_t(c(t))$ from obstacles, and makes progress toward $v_{\text{goal}}$.
\end{problem}

\noindent 
The key difference is that Problem~\ref{pb:global} aims to compute an optimal long-horizon path offline using a fixed calibration dataset, whereas Problem~\ref{pb:local} focuses on finding a feasible path online that is calibration dataset free
and adapts to distribution shifts detected through real-time feedback.

\section{Solution}
We present two planning frameworks that address uncertainty in dynamic environments through conformal prediction. The first framework (Section~\ref{sec:global}) solves Problem~\ref{pb:global} for global, long-horizon planning using pre-calibrated conformal dataset. We develop two algorithmic approaches: a space-time planning formulation with explicit confidence enumeration, and a computationally efficient Safe Interval Path Planning (SIPP) extension that compresses temporal information while maintaining probabilistic safety guarantees. The second framework (Section~\ref{sec:local}) solves Problem~\ref{pb:local} for local reactive planning with online observations, where we develop a sampling-based approach that adaptively calibrates safety bounds in response to distribution shift.

\subsection{Global Time-Aware Planning}
\label{sec:global}

\textbf{3.1.1. Space-time Planning with Confidence:}
In traditional space-time planning, the dynamic environment is modeled as a graph where states $s = (v, t)$ explicitly combine spatial locations $v \in V$ with discrete time steps $t \in \{0, 1, \ldots, T\}$. This formulation produces a state space of size $O(|V| \times T)$ by enumerating all spatial-temporal configurations. While this approach requires exploring a large state space, it guarantees finding the globally optimal solution to Problem~\ref{pb:global}.

To integrate conformal prediction with space-time planning, we augment the space-time representation to include confidence levels: $\hat{s} = (v, c, t)$,
where $v \in V$ is the spatial vertex, $c \in \mathcal{C}$ is the discrete confidence level from our finite confidence set $\mathcal{C} = \{c_1, c_2, \ldots, c_m\}$, and $t \in \{0, 1, \ldots, T\}$ is the time step.
We construct a confidence space-time graph $G_{st} = (V_{st}, E_{st})$ where:
\begin{itemize}[left=0pt]
    \item $V_{st} = \{(v,c,t) : v \in V, t \in \{0, \ldots, T\}, c \in \mathcal{C}, c(v,t) \geq c_{min}\}$ where for each state, $c$ represents the discrete confidence level assigned based on the empirical confidence $c(v,t)$.
    \item $E_{st} = \{((v,c,t), (v',c',t')) : (v,v') \in E, t' = t + w(v,v'), c' \text{ is the confidence level at } (v',t')\}$.
\end{itemize}

Given the confidence space-time graph $G_{ST}$, we find optimal paths using standard graph search algorithms such as A*. Edge weights combine travel time and safety:
\begin{equation}
w((v,c,t), (v',c',t')) = (1-\gamma) \cdot w(v,v') - \gamma \cdot \log(c')
\end{equation}
where $\gamma \in [0,1]$ controls the trade-off as in~\eqref{eq:objective}: $\gamma = 0$ yields purely time-optimal paths, $\gamma = 1$ maximizes confidence, and intermediate values balance both objectives. When $\gamma = 0$ and multiple paths achieve equal travel time, ties are broken by preferring higher-confidence states.

The key advantage of this approach is its completeness: the explicit enumeration of all spatial-temporal-confidence combinations guarantees finding the globally optimal solution. However, this comes at the cost of a state space that grows as $O(|V| \times T)$, which becomes prohibitive for long planning horizons. This motivates our confidence-augmented SIPP approach in the next section, which achieves similar optimality guarantees with dramatically reduced computational complexity.



\textbf{3.1.2. CP-Augmented SIPP: }
While the space--time formulation can achieve globally optimal solutions by explicitly enumerating all spatiotemporal states, it becomes computationally prohibitive as planning horizons or environmental complexity increase. The number of discrete time steps grows rapidly, leading to an explosion in state--time combinations and making direct search intractable for large-scale problems.
Safe Interval Path Planning ~\citep{phillips2011sipp} mitigates this issue by recognizing that, although time is continuous, the number of \emph{contiguous safe intervals} at each location is typically much smaller than the total number of timesteps. Instead of maintaining one state per time step, SIPP aggregates all collision-free times at a vertex into maximal intervals, thereby compressing temporal information without sacrificing optimality with respect to arrival time.
In the standard SIPP formulation, each spatial vertex $v \in V$ maintains a timeline of alternating safe and unsafe intervals. 
A safe interval $I = [t_a, t_b)$ denotes one maximal contiguous period during which vertex $v$ is guaranteed to be collision-free.

Each state is represented as $s = (v, I)$, indicating that the agent is at vertex $v$ during interval $I$. 
Thus, a single vertex may correspond to multiple non-overlapping safe intervals, each forming a distinct state.
This representation dramatically reduces the number of states compared with the full space--time grid. 
During search, SIPP stores for each state $s$ the earliest feasible arrival time $g(s) \in [t_a, t_b]$, which serves as the accumulated cost (i.e., the $g$-value in A*). 
The heuristic $h(s)$ estimates the remaining travel time to the goal (e.g., Euclidean distance divided by maximum speed), and the evaluation function is
$f(s) = g(s) + h(s).$

When expanding a state, SIPP determines the earliest departure time within $I$ that allows collision-free arrival at a successor state. 
This on-the-fly computation of feasible transitions based on arrival time is what enables SIPP’s temporal compression and efficiency.
To incorporate probabilistic safety guarantees, we extend SIPP by discrete confidence levels 
$\mathcal{C} = \{c^1, c^2, \ldots, c^m\}$ in decreasing order, where each $c^i \in [0,1]$ represents a required confidence threshold. 
A confidence-augmented state is defined as $\hat{s} = (v, c, I)$,
where $v \in V$ is a spatial vertex, $c \in \mathcal{C}$ is the confidence level, and $I = [t_a, t_b)$ is the safe interval corresponding to that confidence. 
For each vertex--confidence pair $(v, c)$, we compute the corresponding safe time intervals where vertex $v$ remains collision-free at level $c$:
$\min_{i \in \{1,\ldots,n\}} \|v - \tau_i(t)\| > Q_{1-\alpha}^c(t),
\quad \forall t \in [t_a, t_b),$
where $Q_{1-\alpha}^c(t)$ is the $(1-\alpha)$-quantile of the prediction error from the conformal prediction model. 
In practice, safe intervals are obtained by discretizing time and identifying the maximal contiguous segments satisfying the above condition.

The search cost is defined in terms of time,
$\hat{g}(\hat{s}) = t_{\mathrm{arrival}},$
where $t_{\mathrm{arrival}}$ denotes the arrival time at state $\hat{s}$.
The heuristic function is given by
$\hat{h}(\hat{s}) = h(v, v_{\mathrm{goal}}),$
representing the estimated remaining travel time from vertex $v$ to the goal vertex $v_{\mathrm{goal}}$.
When multiple states share the same $f$-score, the algorithm prioritizes those with higher confidence levels, thereby biasing the search toward safer trajectories without sacrificing time optimality.
If both the $f$-scores and confidence levels are identical, ties are resolved arbitrarily.
Each state transition $(v_j, c_j, I_j, t_j) \rightarrow (v_{j+1}, c_{j+1}, I_{j+1}, t_{j+1})$ must satisfy:
(i). spatial connectivity: $(v_j, v_{j+1}) \in E$;
(ii). temporal feasibility: $t_{j+1} = t_j + w(v_j, v_{j+1})$, where $w(\cdot)$ denotes the travel time;
(iii). departure constraint: $t_j \le t_{\mathrm{end}}^{I_j}$; and
(iv) arrival constraint: $t_{\mathrm{start}}^{I_{j+1}} \le t_{j+1} \le t_{\mathrm{end}}^{I_{j+1}}$.
The confidence level $c_{j+1}$ is selected from any admissible value at vertex $v_{j+1}$ satisfying $c_{j+1} \ge c_{\min}$.
Successor states below this threshold are pruned to improve efficiency.

\begin{theorem}[Trajectory-Level Safety under Marginal Conformal Guarantees]
 \label{thm:traj_safety_sipp}
Let $\pi$ be any trajectory with collision-check grid $\mathcal{T}(\pi) = \{t_0, \ldots, t_k\}$ generated by a planner employing conformal prediction. 
For each $t \in \mathcal{T}(\pi)$, let $E_t$ denote the event that the robot’s configuration at time $t$ lies within the conformal safety set $\mathcal{S}^c_t$ constructed at confidence level $c_t \in [0,1]$. 
Assume that the per-time coverage guarantees hold marginally, that is, $\Pr(E_t) \ge c_t$ for all $t \in \mathcal{T}(\pi)$.

Then, the probability that the trajectory remains safe at all times satisfies
$\Pr\!\Big(\bigcap_{t \in \mathcal{T}(\pi)} E_t\Big)
\ge 1 - \sum_{t \in \mathcal{T}(\pi)} (1 - c_t),$
Equivalently, the probability of a collision or safety violation along the trajectory satisfies
$\Pr\!\Big(\bigcup_{t \in \mathcal{T}(\pi)} E_t^{\complement}\Big)
\le \sum_{t \in \mathcal{T}(\pi)} (1 - c_t),$
where $E_t^{\complement}$ denotes the complement of $E_t$. 
In particular, if the confidence level is uniform, $c_t \equiv 1 - \alpha$, then 
$\Pr\!\Big(\bigcup_{t \in \mathcal{T}(\pi)} E_t^{\complement}\Big) \le k \alpha.$
\end{theorem}


\textbf{Proof}
Let $E_t$ be the per-time safety events with marginal coverage $\Pr(E_t)\ge c_t$. By the complement identity, $\Pr\!\Big(\bigcap_{t\in\mathcal{T}(\pi)} E_t\Big)
= 1 - \Pr\!\Big(\bigcup_{t\in\mathcal{T}(\pi)} E_t^{\complement}\Big)$. Boole's inequality state that $\Pr(\cup_t A_t)\le \sum_t \Pr(A_t)$ for any
events $\{A_t\}$~\citep{10.1093/oso/9780198572237.001.0001}. Applying it with $A_t=E_t^{\complement}$ and using $\Pr(E_t^{\complement})=1-\Pr(E_t)\le 1-c_t$ yields $\Pr\!\Big(\bigcup_{t\in\mathcal{T}(\pi)} E_t^{\complement}\Big)
\le \sum_{t\in\mathcal{T}(\pi)} (1-c_t)$ hence $\Pr(\cap_{t\in\mathcal{T}(\pi)} E_t)\ge 1-\sum_{t\in\mathcal{T}(\pi)}(1-c_t)$. In the uniform case $c_t\equiv 1-\alpha$, this reduces to $\Pr(\cup_{t\in\mathcal{T}(\pi)} E_t^{\complement})\le k\alpha$ with $k=|\mathcal{T}(\pi)|$.

\subsection{Local Reactive Planning with Adaptive Conformal Prediction}
\label{sec:local}
While the confidence-augmented SIPP formulation provides a principled and complete solution for time-aware motion planning on discretized state spaces, its reliance on a pre-defined graph and enumerated safe intervals limits scalability in high-dimensional or continuous domains. To address this, we extend the same conformal-safety framework to a sampling-based setting, where the planner incrementally explores the continuous space–time manifold using random sampling rather than explicit graph expansion. In this regime, the safety of each candidate motion is evaluated through adaptive conformal prediction (ACP), which supplies calibrated uncertainty bounds on obstacle trajectories for a time-varying confidence schedule. This yields a time-aware conformal RRT (ACP-RRT) that preserves the probabilistic safety guarantees of the SIPP formulation while offering the flexibility and scalability of sampling-based planning.

\subsubsection{Calibration-free Adaptive Conformal Prediction}

\textbf{Exchangeability Diagnostics:}
Real deployments rarely preserve exchangeability between the historical calibration dataset and the incoming prediction-observation stream. Before passing confidence to SIPP, we therefore insert an exchangeability gate.Let $\mathcal D_{\mathrm{cal}}=\{R_j\}_{j=1}^{n}$ be the calibration scores defined in \S\ref{sec:cp-basics}. As soon as feedback becomes availabel online, we collect a warm-up batch of new scores $\mathcal D_{\mathrm{new}}=\{R_t\}_{t=t_0}^{t_0+W_0-1}$. We assess wether the marginal score distribution remains unchange across the $\mathcal D_{\mathrm{cal}}$ and $\mathcal D_{\mathrm{new}}$. The equality condition is $\mathcal L(R\mid \mathrm{cal})=\mathcal L(R\mid \mathrm{new})$.

Deviation from this criterion is quantified by the two-sample Kolmogorov–Smirnov (KS) distance $D=\sup_{x\in\mathbb R}\big|\hat F_{\mathrm{cal}}(x)-\hat F_{\mathrm{new}}(x)\big|$, where $\hat F_{\mathrm{cal}}$ and $\hat F_{\mathrm{new}}$ are the empirical CDFs of $\mathcal D_{\mathrm{cal}}$ and $\mathcal D_{\mathrm{new}}$. To maintain the nominal level under short-range serial dependence, we approximate the null distribution of $D$ using a time-robust permutation: pool $Z=(\mathcal D_{\mathrm{cal}}\Vert \mathcal D_{\mathrm{new}})$, split $Z$ into contiguous blocks of length $B$, randomly permute the blocks to obtain $Z^\ast$, and resplit $Z^\ast$ into $(\mathcal D_{\mathrm{cal}}^\ast,\mathcal D_{\mathrm{new}}^\ast)$. The permuted statistic is 
\begin{equation}
D^\ast=\sup_{x\in\mathbb R}\big|\hat F_{\mathrm{cal}}^\ast(x)-\hat F_{\mathrm{new}}^\ast(x)\big|
\end{equation}

Repeating this procedure $N$ times yields $\{D^{\ast(b)}\}_{b=1}^{N}$ and the unbiased permutation p-value
\begin{equation}
p=\frac{1+\sum_{b=1}^{N}\mathbf 1\!\left\{D^{\ast(b)}\ge D\right\}}{N+1}
\end{equation}
if $p<\alpha_{\mathrm{gate}}$, the gate rejects and ACP is activated.

\textbf{Adaptive update rule} When the exchangeability test rejects, we discontinue the use of calibration quantiles and switch to calibration-free ACP. Let $R_t$ be the nonconformity score at time $t$. We introduce a positive scale $\lambda_t$ and a fixed monotone threshold map $C(\lambda)$.
\begin{equation}
H(\lambda_t)=\lambda_t\, d_{\min}(s_t),\qquad
d_{\min}(s_t)=\min_{i}\ \inf_{\tau\in I(s_t)} \big\|v(s_t)-\hat r_i(\tau)\big\|
\end{equation}

where $s_t$ is the current vertex-time state, $I(s_t)$ is its SIPP safe interval, $v(s_t)$ is the spatial location at $s_t$,and $\hat r_i(\cdot)$ denotes the predicted trajectory of obstacle $i$. Define the miscoverage indicator
\begin{equation}
e_t=\mathbf{1}\{R_t >   H_t(\lambda_t)\}\in\{0,1\}
\end{equation}

During deployment the map $H(\cdot)$ remains fixed; only the positive scale $\lambda_t$ adapts online, we update
\begin{equation}
\lambda_{t+1}=\Pi_{[\lambda_{\min},\lambda_{\max}]}(\lambda_t\exp\{\kappa(e_t-\alpha)\})
\end{equation}
initialized at $\lambda_0=1$. Under miscoverage ($e_t=1$) increases $\lambda_t$ so that the next region is more conservative; under coverage ($e_t=0$) it decreases. 

\textbf{Regularity} We assume that the miscoverage response $p(\lambda)=\mathbb E[e_t\mid \lambda_t=\lambda,\mathcal F_{t-1}]$ is strictly decreasing in $\lambda$ on a compact interval $[\lambda_{\min},\lambda_{\max}]$ that brackets the target, $p(\lambda_{\min})>\alpha$ and $p(\lambda_{\max})<\alpha$. Feedback exhibits short-range dependence so that time averages stabilize. Under these standard conditions and a small constant step size, the one-dimensional feedback on $\lambda_t$ yields the time-average tracking guarantee state below.


\begin{theorem}[Time-average coverage control for calibration-free ACP]
Under the regularity ab-ove and a constant step size $\kappa>0$, the closed loop remains in  $[\lambda_{\min},\lambda_{\max}]$ and the realized miscoverage frequency satisfies
\[\Bigg|\frac{1}{T}\sum_{t=1}^{T} e_t - \alpha\Bigg|\;=\; \mathcal{O}_{\mathbb{P}}(\kappa)\;+\;\mathcal{O}_{\mathbb{P}}(T^{-1/2})\] Therefore a sufficiently small $\kappa$ keeps the time-average miscoverage in an $\mathcal{O}_\kappa$ band around $\alpha$ without assuming exchangeability with the historical data. 
\end{theorem}

\textbf{Proof:} Let $z_t=\log\lambda_t$, so the multiplicative update becomes the projected additive recursion $z_{t+1}=\Pi_{[\log\lambda_{\min},\,\log\lambda_{\max}]}\!\bigl(z_t+\kappa(e_t-\alpha)\bigr)$. Write $e_t=p(\lambda_t)+\xi_t$, $\{\xi_t\}$ is bounded martingale difference under the same weak dependence used by the gate. Let $z^\star=\log\lambda^\star$ with $p(\lambda^\star)=\alpha$, and set $V_t=(z_t-z^\star)^2$. By strict monotonicity of $p$ in the compact interval $[\lambda_{\min},\lambda_{\max}]$, for every $\varepsilon>0$ there exists $\rho(\varepsilon)>0$ such that whenever $|z_t-z^\star|\ge \varepsilon$ one has $(z_t-z^\star)\{p(\lambda_t)-\alpha\} \;\le\; -\,\rho(\varepsilon)\,(z_t-z^\star)^2$.
Using projection nonexpansiveness, $\mathbb E[V_{t+1}-V_t\mid\mathcal F_{t-1}] \le -2\kappa\rho(\varepsilon)\,V_t+\kappa^2$. Summing yields $\frac1T\sum_{t=1}^T \mathbb E[V_t]=O(\kappa)$. Averaging $e_t-\alpha=\{p(\lambda_t)-\alpha\}+\xi_t$ gives $\Big|\tfrac1T\sum_{t=1}^T (e_t-\alpha)\Big| = O(\kappa)+\mathcal O_{\mathbb P}(T^{-1/2})$ Since $\tfrac1T\sum\xi_t=\mathcal O_{\mathbb P}(T^{-1/2})$. With diminishing step sizes satisfying $\sum_t\kappa_t=\infty$ and $\sum_t\kappa_t^2<\infty$, the bias vanishes and $\tfrac1T\sum_{t=1}^T e_t \xrightarrow{\mathbb P}\alpha$.

\subsubsection{time-aware conformal RRT}
The time-aware conformal RRT extends space-time RRT approaches~\citep{grothe2022st,sintov2014time}, 
which represent nodes as $(x, t)$ pairs, by augmenting each node with a confidence 
level $c$ to explicitly reason about prediction uncertainty through conformal prediction.
When expanding the tree, the algorithm samples a random spatial state, connects it to the nearest node, and assigns an arrival time based on the travel distance and robot speed. Along each edge, the planner predicts the robot's motion forward in time and checks for collisions against predicted obstacle trajectories. At each intermediate step, the confidence level \(c(t)\) is obtained from a predefined decay schedule, and ACP provides the corresponding prediction radius for that horizon. The edge is accepted only if all sampled points along it remain outside the ACP-inflated obstacle regions, ensuring time-consistent conformal safety.

To avoid the problem being infeasible due to a long prediction horizon and to capture the increasing uncertainty of long-term forecasts, the planner employs a time-varying confidence schedule \(c(t)\) that smoothly decays from an initial high confidence \(c_{\text{start}} \in \mathcal{C}\) to a lower terminal confidence \(c_{\text{end}} \in \mathcal{C}\) over the planning horizon \(H\):

\begin{equation}
\label{eq: adaptive_conf}
    c(t) = c_{\text{end}} + (c_{\text{start}} - c_{\text{end}})\left(1 - \frac{t}{H}\right).
\end{equation}
This schedule reflects the intuition that near-term predictions are more reliable and should be treated conservatively, while distant predictions can tolerate greater uncertainty. As summarized in Algorithm~\ref{alg:acp-rrt}, the planner grows the conformal RRT in a receding-horizon manner: at each planning cycle, it constructs a time-aware tree using the current ACP calibration based on prediction step and requiring confidence, executes only the first motion segment (RRT node), and then incorporates new observations to update the ACP bounds. This closed-loop procedure continuously adapts both the uncertainty estimates and the effective confidence level over time.

\begin{algorithm}[hbt!]
\SetAlgoNoEnd
\DontPrintSemicolon
\SetAlgoLined
\LinesNumbered
\caption{ACP-RRT}
\label{alg:acp-rrt}
\KwIn{Start node $(x_s, t_s, c_s)$, goal region $\mathcal{X}_{\text{goal}}$, ACP model $\mathcal{A}$}
\KwOut{Probabilistically safe path to $\mathcal{X}_{\text{goal}}$}

$\mathcal{T} \leftarrow \{(x_s, t_s, c_s)\}$\;

\While{goal not reached}{
    Sample $x_{\text{rand}} \in \mathcal{X}$\;
    $(x_{\text{near}}, t_{\text{near}}, c_{\text{near}}) \leftarrow \text{Nearest}(\mathcal{T}, x_{\text{rand}})$\;\\
    $x_{\text{new}} \leftarrow \text{Steer}(x_{\text{near}}, x_{\text{rand}})$\;\\
    $t_{\text{new}} \leftarrow t_{\text{near}} + \frac{\|x_{\text{new}} - x_{\text{near}}\|}{v_{\max}}$\;\\
    $c_{\text{new}} \leftarrow c(t_{\text{new}})$\ \text{from}~\eqref{eq: adaptive_conf};\\
    \If{$|x_{\text{new}} - \tau_i| > \text{ACP}(t_{\text{new}}, c_{\text{new}})$ for all $i$}{
        Add $(x_{\text{new}}, t_{\text{new}}, c_{\text{new}})$ to $\mathcal{T}$\;
        \If{$x_{\text{new}} \in \mathcal{X}_{\text{goal}}$}{\Return{Path($\mathcal{T}$)}}
    }
}
\Return{failure}\;
\end{algorithm}

\section{Results}
\label{sec:results}
We evaluate our two planning frameworks through simulation experiments. Section~\ref{sec:results_sipp} presents results for CP-SIPP, demonstrating global planning performance on grid environments with dynamic obstacles. Section~\ref{sec:results_rrt} evaluates ACP-RRT for local reactive planning in continuous domains, showing how adaptive conformal prediction maintains safety under distribution shift. We analyze path quality, and safety guarantees for both approaches.

\subsection{CP-SIPP}
\label{sec:results_sipp}
\begin{figure}[hbt!]
    \centering
    \includegraphics[width=1\linewidth]{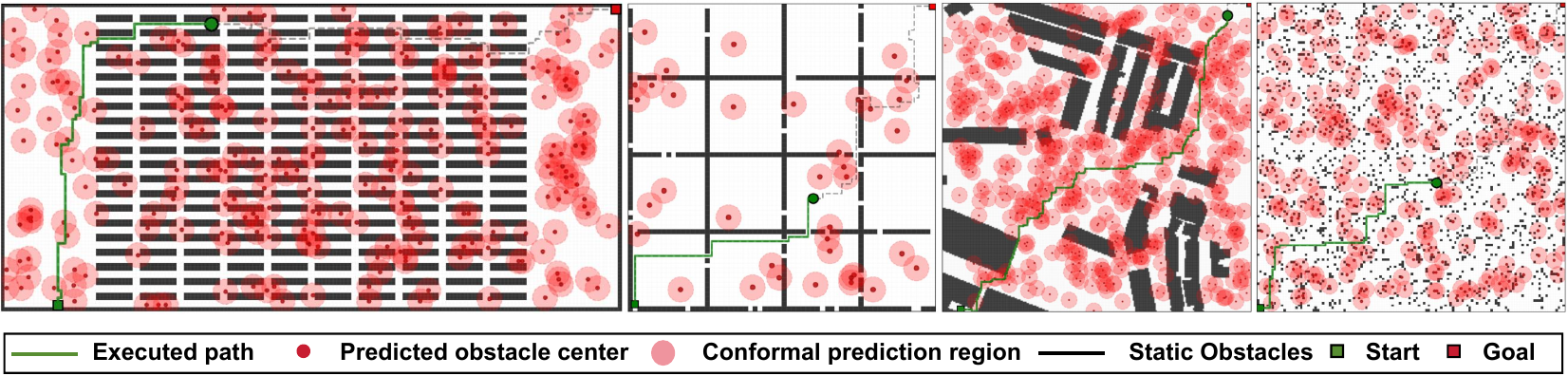}
    \caption{CP-SIPP: four frames showing uncertainty-aware CP-SIPP navigating through complex environments with dynamic obstacles}
    \label{fig:astar_path}
\end{figure}

To further illustrate how uncertainty evolves across prediction horizons and confidence levels, 
we visualize the quantile table generated by the Seq2Seq LSTM-based motion prediction model~\citep{sutskever2014sequencesequencelearningneural}. 
As shown in Fig.~\ref{fig:astar_path}, CP-SIPP demonstrates its ability to perform uncertainty-aware navigation in complex environments with dynamic obstacles. By incorporating conformal prediction into the SIPP framework, the planner adaptively adjusts its trajectory to avoid predicted obstacle regions while maintaining computational efficiency. The green line indicates the executed path, dark red dots represent predicted obstacle centers, and light red disks denote their 95\% quantile conformal prediction confidence regions rendered at the current timestep.
\subsection{ACP-RRT}
\label{sec:results_rrt}

\begin{wrapfigure}[7]{r}{0.5\textwidth}  
    \vspace{-70pt}  
    \centering    \includegraphics[width=0.48\textwidth]{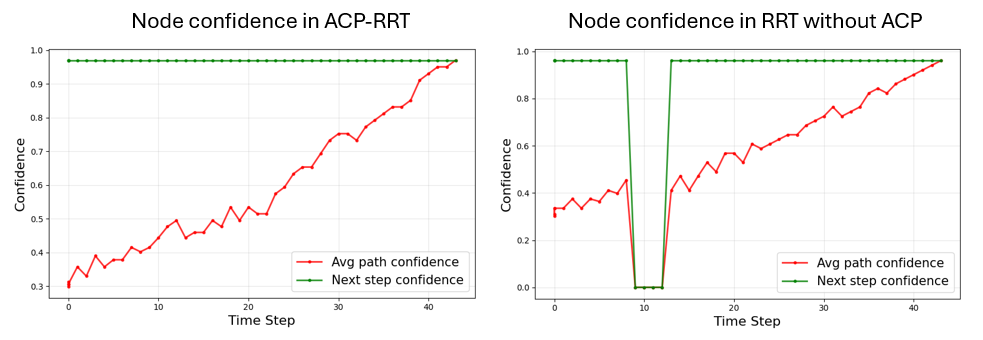}
    \captionsetup{format=plain, singlelinecheck=false}
    \caption{Confidence Evolution Comparison: left: The ACP-RRT planner successfully maintains high confidence. right: The baseline RRT 
    (without ACP) fails with a collision at $t=[9,12]$.}
    \label{fig:avg_path}
\end{wrapfigure}

\begin{figure}[hbt!]
    \centering
    \includegraphics[width=1\linewidth]{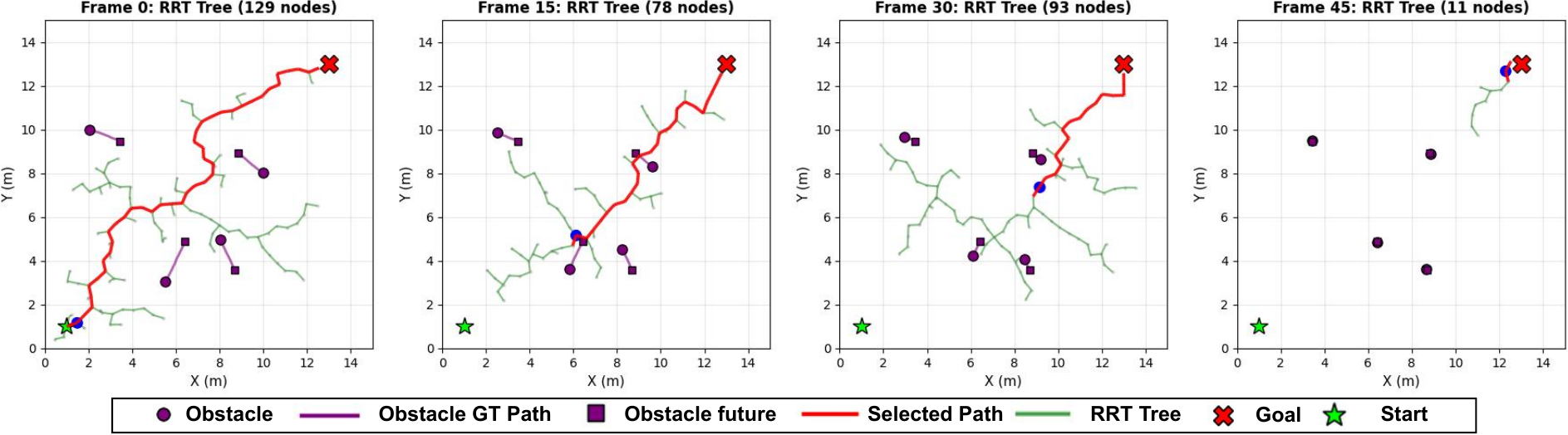}
    \caption{ACP-RRT: three frames showing online ACP-RRT adaptive generating trees to navigating through obstacles with uncertainty}
    \label{fig:rrt_path}
\end{figure}

Fig.~\ref{fig:rrt_path} illustrates the performance of the proposed ACP-RRT with a planning horizon of $H = 50$. We introduce three dynamic obstacles, and ACP is employed to generate confidence regions for their predicted 
trajectories at each time step. The figure presents snapshots of the planning process at $t = 0$, $t = 25$, and $t = 45$. As the prediction horizon increases, the RRT accepts nodes with lower confidence levels, allowing it to maintain feasibility while adapting to the growing trajectory prediction uncertainty.


Fig.~\ref{fig:avg_path} (left) illustrates the evolution of the average node confidence during planning. As time progresses, the agent requires fewer steps to reach the target, resulting in shorter prediction horizons and consequently requiring fewer confidence regions. This leads to an overall increase in the average node confidence of the constructed RRT path over time. Meanwhile, due to the receding-horizon structure and the adaptive confidence update rule in~\eqref{eq: adaptive_conf}, the confidence associated with the immediate next step consistently remains the highest throughout the planning process. 
In contrast, Fig.~\ref{fig:avg_path} (right) shows the baseline method 
implementing reactive RRT without ACP, where confidence is validated 
post-sampling rather than proactively incorporated during tree expansion. 
This approach fails with a collision at $t=[9, 12]$, where the confidence 
level drops to zero.

\section{Conclusion}
We presented two motion planning frameworks that integrate conformal prediction to enable uncertainty-aware navigation in dynamic environments. The first, CP-SIPP, extends the classical SIPP formulation by incorporating discrete confidence levels derived from conformal prediction, allowing the planner to compute time-optimal trajectories under formal, distribution-free probabilistic safety guarantees. The second, ACP-RRT, generalizes these principles to continuous domains through a sampling-based approach that adaptively adjusts safety bounds online. Together, these frameworks demonstrate how conformal prediction can provide a unified foundation for efficient, and provably safe motion planning under uncertainty.

\bibliography{l4dc2026-sample}

@article{lindemann2023safe,
  title={Safe planning in dynamic environments using conformal prediction},
  author={Lindemann, Lars and Cleaveland, Matthew and Shim, Gihyun and Pappas, George J},
  journal={IEEE Robotics and Automation Letters},
  volume={8},
  number={8},
  pages={5116--5123},
  year={2023},
  publisher={IEEE}
}

@article{yu2026signal,
  title={Signal temporal logic control synthesis among uncontrollable dynamic agents with conformal prediction},
  author={Yu, Xinyi and Zhao, Yiqi and Yin, Xiang and Lindemann, Lars},
  journal={Automatica},
  volume={183},
  pages={112616},
  year={2026},
  publisher={Elsevier}
}

@article{wang2024conformal,
  title={Conformal temporal logic planning using large language models},
  author={Wang, Jun and Tong, Jiaming and Tan, Kaiyuan and Vorobeychik, Yevgeniy and Kantaros, Yiannis},
  journal={ACM Transactions on Cyber-Physical Systems},
  year={2024},
  publisher={ACM New York, NY}
}

@inproceedings{chee2024uncertainty,
  title={Uncertainty quantification and robustification of model-based controllers using conformal prediction},
  author={Chee, Kong Yao and Silva, Thales C and Hsieh, M Ani and Pappas, George J},
  booktitle={6th Annual Learning for Dynamics \& Control Conference},
  pages={528--540},
  year={2024},
  organization={PMLR}
}

@inproceedings{sintov2014time,
  title={Time-based RRT algorithm for rendezvous planning of two dynamic systems},
  author={Sintov, Avishai and Shapiro, Amir},
  booktitle={2014 IEEE International Conference on Robotics and Automation (ICRA)},
  pages={6745--6750},
  year={2014},
  organization={IEEE}
}

@inproceedings{grothe2022st,
  title={St-rrt*: Asymptotically-optimal bidirectional motion planning through space-time},
  author={Grothe, Francesco and Hartmann, Valentin N and Orthey, Andreas and Toussaint, Marc},
  booktitle={2022 International Conference on Robotics and Automation (ICRA)},
  pages={3314--3320},
  year={2022},
  organization={IEEE}
}

@article{garone2017reference,
  title={Reference and command governors for systems with constraints: A survey on theory and applications},
  author={Garone, Emanuele and Di Cairano, Stefano and Kolmanovsky, Ilya},
  journal={Automatica},
  volume={75},
  pages={306--328},
  year={2017},
  publisher={Elsevier}
}

@inproceedings{liang2024control,
  title={Control Barrier Function for Linearizable Systems with High Relative Degrees from Signal Temporal Logics: A Reference Governor Approach},
  author={Liang, Kaier and Cai, Mingyu and Vasile, Cristian-Ioan},
  booktitle={2024 American Control Conference (ACC)},
  pages={1676--1681},
  year={2024},
  organization={IEEE}
}

@inproceedings{ames2019control,
  title={Control barrier functions: Theory and applications},
  author={Ames, Aaron D and Coogan, Samuel and Egerstedt, Magnus and Notomista, Gennaro and Sreenath, Koushil and Tabuada, Paulo},
  booktitle={2019 18th European control conference (ECC)},
  pages={3420--3431},
  year={2019},
  organization={Ieee}
}

@article{lopez2020robust,
  title={Robust adaptive control barrier functions: An adaptive and data-driven approach to safety},
  author={Lopez, Brett T and Slotine, Jean-Jacques E and How, Jonathan P},
  journal={IEEE Control Systems Letters},
  volume={5},
  number={3},
  pages={1031--1036},
  year={2020},
  publisher={IEEE}
}

@article{barber2023conformal,
  title={Conformal prediction beyond exchangeability},
  author={Barber, Rina Foygel and Candes, Emmanuel J and Ramdas, Aaditya and Tibshirani, Ryan J},
  journal={The Annals of Statistics},
  volume={51},
  number={2},
  pages={816--845},
  year={2023},
  publisher={Institute of Mathematical Statistics}
}

@inproceedings{mossina2024conformal,
  title={Conformal semantic image segmentation: Post-hoc quantification of predictive uncertainty},
  author={Mossina, Luca and Dalmau, Joseba and And{\'e}ol, L{\'e}o},
  booktitle={Proceedings of the IEEE/CVF Conference on Computer Vision and Pattern Recognition},
  pages={3574--3584},
  year={2024}
}

@article{cherian2024large,
  title={Large language model validity via enhanced conformal prediction methods},
  author={Cherian, John and Gibbs, Isaac and Candes, Emmanuel},
  journal={Advances in Neural Information Processing Systems},
  volume={37},
  pages={114812--114842},
  year={2024}
}

@book{vovk2005algorithmic,
  title={Algorithmic learning in a random world},
  author={Vovk, Vladimir and Gammerman, Alexander and Shafer, Glenn},
  year={2005},
  publisher={Springer}
}

@inproceedings{phillips2011sipp,
  title={Sipp: Safe interval path planning for dynamic environments},
  author={Phillips, Mike and Likhachev, Maxim},
  booktitle={2011 IEEE international conference on robotics and automation},
  pages={5628--5635},
  year={2011},
  organization={IEEE}
}

@article{chow2015risk,
  title={Risk-sensitive and robust decision-making: a cvar optimization approach},
  author={Chow, Yinlam and Tamar, Aviv and Mannor, Shie and Pavone, Marco},
  journal={Advances in neural information processing systems},
  volume={28},
  year={2015}
}

@article{zhang2022robust,
  title={Robust tube-based model predictive control with Koopman operators},
  author={Zhang, Xinglong and Pan, Wei and Scattolini, Riccardo and Yu, Shuyou and Xu, Xin},
  journal={Automatica},
  volume={137},
  pages={110114},
  year={2022},
  publisher={Elsevier}
}

@article{du2011robot,
  title={Robot motion planning in dynamic, uncertain environments},
  author={Du Toit, Noel E and Burdick, Joel W},
  journal={IEEE Transactions on Robotics},
  volume={28},
  number={1},
  pages={101--115},
  year={2011},
  publisher={IEEE}
}

@article{mayne2011tube,
  title={Tube-based robust nonlinear model predictive control},
  author={Mayne, David Q and Kerrigan, Erric C and Van Wyk, EJ and Falugi, Paola},
  journal={International journal of robust and nonlinear control},
  volume={21},
  number={11},
  pages={1341--1353},
  year={2011},
  publisher={Wiley Online Library}
}

@article{hakobyan2021wasserstein,
  title={Wasserstein distributionally robust motion control for collision avoidance using conditional value-at-risk},
  author={Hakobyan, Astghik and Yang, Insoon},
  journal={IEEE Transactions on Robotics},
  volume={38},
  number={2},
  pages={939--957},
  year={2021},
  publisher={IEEE}
}

@article{blackmore2011chance,
  title={Chance-constrained optimal path planning with obstacles},
  author={Blackmore, Lars and Ono, Masahiro and Williams, Brian C},
  journal={IEEE Transactions on Robotics},
  volume={27},
  number={6},
  pages={1080--1094},
  year={2011},
  publisher={IEEE}
}

@article{strawn2023conformal,
  title={Conformal predictive safety filter for rl controllers in dynamic environments},
  author={Strawn, Kegan J and Ayanian, Nora and Lindemann, Lars},
  journal={IEEE Robotics and Automation Letters},
  volume={8},
  number={11},
  pages={7833--7840},
  year={2023},
  publisher={IEEE}
}

@article{sun2023conformal,
  title={Conformal prediction for uncertainty-aware planning with diffusion dynamics model},
  author={Sun, Jiankai and Jiang, Yiqi and Qiu, Jianing and Nobel, Parth and Kochenderfer, Mykel J and Schwager, Mac},
  journal={Advances in Neural Information Processing Systems},
  volume={36},
  pages={80324--80337},
  year={2023}
}

@article{rudenko2020human,
  title={Human motion trajectory prediction: A survey},
  author={Rudenko, Andrey and Palmieri, Luigi and Herman, Michael and Kitani, Kris M and Gavrila, Dariu M and Arras, Kai O},
  journal={The International Journal of Robotics Research},
  volume={39},
  number={8},
  pages={895--935},
  year={2020},
  publisher={Sage Publications Sage UK: London, England}
}

@article{aoude2013probabilistically,
  title={Probabilistically safe motion planning to avoid dynamic obstacles with uncertain motion patterns},
  author={Aoude, Georges S and Luders, Brandon D and Joseph, Joshua M and Roy, Nicholas and How, Jonathan P},
  journal={Autonomous Robots},
  volume={35},
  number={1},
  pages={51--76},
  year={2013},
  publisher={Springer}
}

@article{mavrogiannis2023core,
  title={Core challenges of social robot navigation: A survey},
  author={Mavrogiannis, Christoforos and Baldini, Francesca and Wang, Allan and Zhao, Dapeng and Trautman, Pete and Steinfeld, Aaron and Oh, Jean},
  journal={ACM Transactions on Human-Robot Interaction},
  volume={12},
  number={3},
  pages={1--39},
  year={2023},
  publisher={ACM New York, NY}
}

@misc{dixit2022adaptiveconformalpredictionmotion,
      title={Adaptive Conformal Prediction for Motion Planning among Dynamic Agents}, 
      author={Anushri Dixit and Lars Lindemann and Skylar Wei and Matthew Cleaveland and George J. Pappas and Joel W. Burdick},
      year={2022},
      eprint={2212.00278},
      archivePrefix={arXiv},
      primaryClass={cs.RO},
}

@misc{gibbs2021adaptiveconformalinferencedistribution,
      title={Adaptive Conformal Inference Under Distribution Shift}, 
      author={Isaac Gibbs and Emmanuel Candès},
      year={2021},
      eprint={2106.00170},
      archivePrefix={arXiv},
      primaryClass={stat.ME},
}

@article{JMLR:v25:22-1218,
  author  = {Isaac Gibbs and Emmanuel J. Cand{{\`e}}s},
  title   = {Conformal Inference for Online Prediction with Arbitrary Distribution Shifts},
  journal = {Journal of Machine Learning Research},
  year    = {2024},
  volume  = {25},
  number  = {162},
  pages   = {1--36},
}

@book{10.1093/oso/9780198572237.001.0001,
    author = {Grimmett, Geoffrey R and Stirzaker, David R},
    title = {Probability and Random Processes},
    publisher = {Oxford University Press},
    year = {2001},
    month = {05},
    isbn = {9780198572237},
    doi = {10.1093/oso/9780198572237.001.0001},
}

@misc{lei2017distributionfreepredictiveinferenceregression,
      title={Distribution-Free Predictive Inference For Regression}, 
      author={Jing Lei and Max G'Sell and Alessandro Rinaldo and Ryan J. Tibshirani and Larry Wasserman},
      year={2017},
      eprint={1604.04173},
      archivePrefix={arXiv},
      primaryClass={stat.ME},
}

@article{liang2025safe,
  title={Safe navigation in dynamic environments using data-driven Koopman operators and conformal prediction},
  author={Liang, Kaier and Yang, Guang and Cai, Mingyu and Vasile, Cristian-Ioan},
  journal={arXiv preprint arXiv:2504.00352},
  year={2025}
}

@misc{sutskever2014sequencesequencelearningneural,
      title={Sequence to Sequence Learning with Neural Networks}, 
      author={Ilya Sutskever and Oriol Vinyals and Quoc V. Le},
      year={2014},
      eprint={1409.3215},
      archivePrefix={arXiv},
      primaryClass={cs.CL},
      url={https://arxiv.org/abs/1409.3215}, 
}


\end{document}